\documentclass[letterpaper]{article} % DO NOT CHANGE THIS
\usepackage{aaai25}  % DO NOT CHANGE THIS
\usepackage{times}  % DO NOT CHANGE THIS
\usepackage{helvet}  % DO NOT CHANGE THIS
\usepackage{courier}  % DO NOT CHANGE THIS
\usepackage[hyphens]{url}  % DO NOT CHANGE THIS
\usepackage{graphicx} % DO NOT CHANGE THIS
\usepackage{natbib}  % DO NOT CHANGE THIS AND DO NOT ADD ANY OPTIONS TO IT
\usepackage{caption} % DO NOT CHANGE THIS AND DO NOT ADD ANY OPTIONS TO IT
\usepackage{algorithm}
\usepackage{algorithmic}
\graphicspath{{imgs/} }
\usepackage{enumitem}
\usepackage{multirow}
\usepackage{amssymb}
\usepackage{booktabs}
\usepackage{amsmath}

\newcommand{\ie}{\textit{i}.\textit{e}.}
\newcommand{\eg}{\textit{e}.\textit{g}.}

\usepackage{newfloat}
\usepackage{listings}
\DeclareCaptionStyle{ruled}{labelfont=normalfont,labelsep=colon,strut=off} % DO NOT CHANGE THIS
\floatstyle{ruled}
\newfloat{listing}{tb}{lst}{}
\floatname{listing}{Listing}
\title{MV-VTON: Multi-View Virtual Try-On with Diffusion Models}
\author{
    Haoyu Wang\textsuperscript{\rm 1,2,}\equalcontrib, Zhilu Zhang\textsuperscript{\rm 2,}\thanks{Corresponding Author.}
    Donglin Di\textsuperscript{\rm 3},
    Shiliang Zhang\textsuperscript{\rm 1},
    Wangmeng Zuo\textsuperscript{\rm 2}
}

\affiliations{
    \textsuperscript{\rm 1}State Key Laboratory of Multimedia Information Processing, School of Computer Science, Peking University\\
    \textsuperscript{\rm 2}Harbin Institute of Technology\\
    \textsuperscript{\rm 3}Space AI, Li Auto\\
    
    cshy2002@gmail.com, cszlzhang@outlook.com,  donglin.ddl@gmail.com, slzhang.jdl@pku.edu.cn, wmzuo@hit.edu.cn
}

\usepackage{bibentry}
\begin{document}

\maketitle

\begin{abstract}
% codes and datasets link
% table data
The goal of image-based virtual try-on is to generate an image of the target person naturally wearing the given clothing. 
However, existing methods solely focus on the frontal try-on using the frontal clothing. 
When the views of the clothing and person are significantly inconsistent, particularly when the person's view is non-frontal, the results are unsatisfactory. 
To address this challenge, we introduce \textbf{M}ulti-\textbf{V}iew \textbf{V}irtual \textbf{T}ry-\textbf{ON} (MV-VTON), which aims to reconstruct the dressing results from multiple views using the given clothes.
Given that single-view clothes provide insufficient information for MV-VTON, we instead employ two images, \ie, the frontal and back views of the clothing, to encompass the complete view as much as possible.
Moreover, we adopt diffusion models that have demonstrated superior abilities to perform our MV-VTON.
In particular, we propose a view-adaptive selection method where hard-selection and soft-selection are applied to the global and local clothing feature extraction, respectively. 
This ensures that the clothing features are roughly fit to the person's view.
Subsequently, we suggest joint attention blocks to align and fuse clothing features with person features.
Additionally, we collect a MV-VTON dataset MVG, in which each person has multiple photos with diverse views and poses.
Experiments show that the proposed method not only achieves state-of-the-art results on MV-VTON task using our MVG dataset, but also has superiority on frontal-view virtual try-on task using VITON-HD and DressCode datasets.
\end{abstract}

\begin{links}
    \link{Code}{https://github.com/hywang2002/MV-VTON}
\end{links}

\begin{figure}[t!]
    \includegraphics[width=0.479\textwidth]{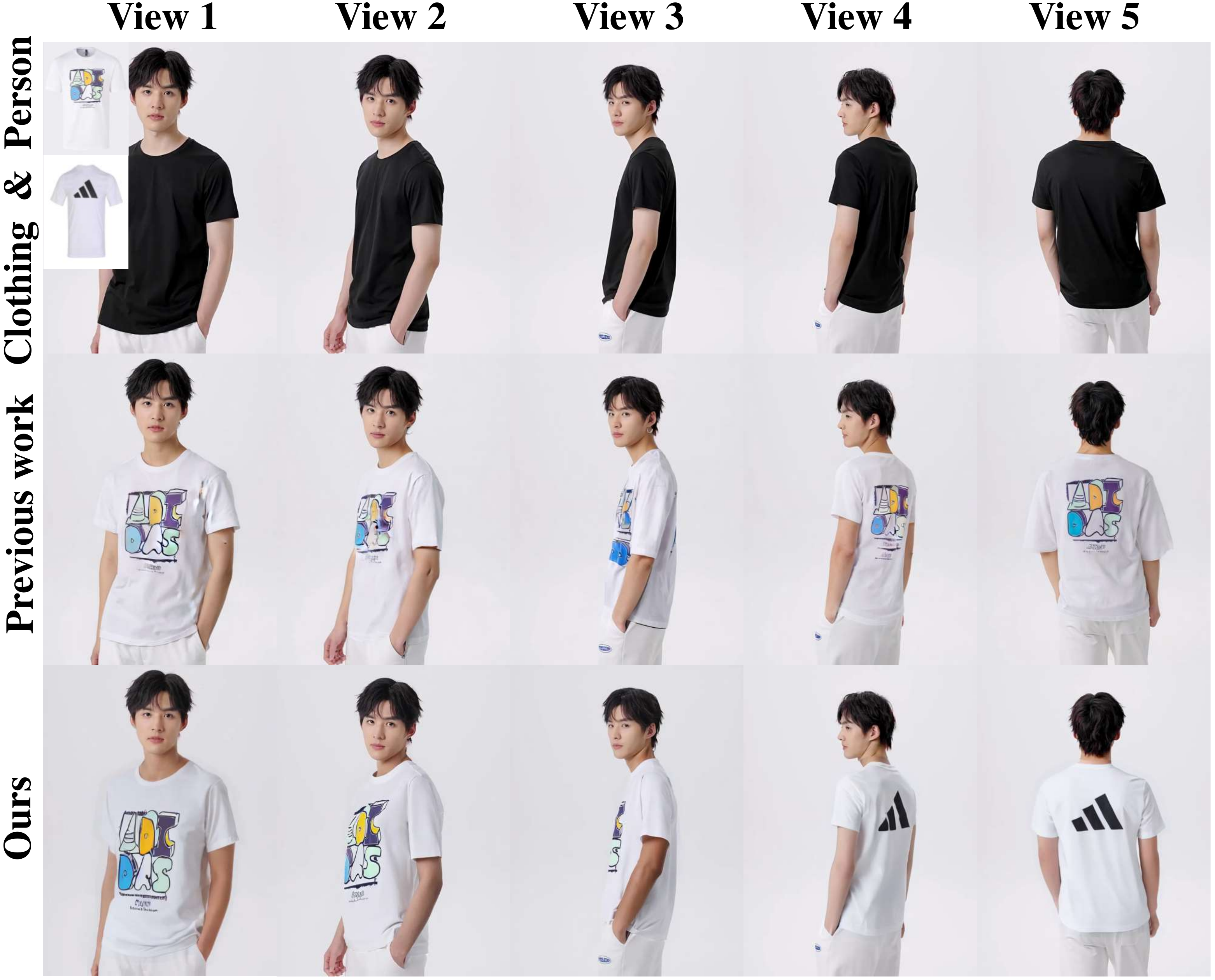}
    \caption{Motivation of this work. Previous VTON methods, \eg, StableVITON~\cite{kim2023stableviton} can only be used for the frontal-view person, and fail when facing the person with multiple views. Our MV-VTON can faithfully present the try-on results for a person with various views.
    }
    \label{fig:motivation}
\end{figure}

\section{Introduction}

Virtual Try-On (VTON) is a classic yet intriguing technology. It can be applied in the field of fashion and clothes online shopping to improve user experience. VTON aims to render the visual effect of a person wearing a specified garment. The emphasis of this technology lies in reconstructing a realistic image that faithfully preserves personal attributes 
% (\eg, appearance and pose)
and accurately represents clothing shape and details.

Early VTON methods~\cite{lee2022high,xie2023gp,bai2022single,he2022style} are based on generative adversarial networks~\cite{goodfellow2020generative} (GANs).
They generally align the clothing to the person's pose, and then employ a generator to fuse the warped clothing with the person.
However, it poses a challenge to ensure that the warped clothing fits the target person's pose, and inaccurate clothes features will easily lead to distortion results. 
% %
Recently, diffusion models~\cite{rombach2022high} have made remarkable strides in the field of image generation~\cite{ruiz2023dreambooth}. Leveraging its potent generative capabilities, some researchers~\cite{morelli2023ladi, kim2023stableviton} have integrated it into virtual try-on fields, building upon previous work and achieving commendable results. 

Although VTON has made great progress, most existing methods focus on performing the frontal try-on.
In practical applications, such as online shopping for clothes, customers may expect to obtain the dressing effect on multiple views (\eg, side or back).
In this case, the pose of the garment may be seriously inconsistent with the person's posture, and the single-view clothing may not be enough to provide complete try-on information.
Thus, these methods will easily generate results with poorly deformed clothing, and lead to the loss of high-frequency details such as texts, patterns, and other textures on clothing, as shown in Figure~\ref{fig:motivation}.

To address these issues, we introduce Multi-View Virtual Try-ON (MV-VTON), which aims to reconstruct the appearance and attire of a person from multiple views. 
%
% MV-VTON should not only address the challenge of natural warping on the clothing’s front but also tackle the pose inconsistency issue between the given clothing and target person. 
%
For example, for clothing in Figure~\ref{fig:motivation}, which may exhibit significant differences between frontal and back styles, MV-VTON should be able to display try-on results in various views, including front, back, and side ones.
Thus, providing single clothing can't meet the needs of dressing up, as the clothing only has partial information. Instead, we utilize both the frontal and back views of the clothing, which covers approximately complete view with as few images as possible.

Given the frontal and back clothing, we utilize the popular diffusion method to achieve MV-VTON.
It is natural but doesn't work well to simply concatenate two pieces of clothing together as conditions of diffusion models, as it is difficult for the model to learn how to assign two-view clothes to a person, especially when the person is sideways. 
Instead, we propose a view-adaptive selection mechanism, which picks appropriate features of two-view clothes based on the posture information of the person and clothes.
Therein, the hard-selection module chooses one of the two clothes for global feature extraction, and the soft-selection module modulates the local features of two clothes.
We utilize CLIP~\cite{radford2021learning} and a multi-scale encoder to extract the global and local clothing features, respectively.
Moreover, to enhance the preservation of high-frequency details in clothing, we present joint attention blocks. They independently align global and local features with the person features, and selectively fuse them to refine the local clothing details while preserving global semantic information.

Furthermore, we collect a multi-view virtual try-on dataset, named Multi-View Garment (MVG). It contains thousands of samples, and each sample contains 5 images under different views and poses.
We conduct extensive experiments not only on MV-VTON task using the MVG dataset, but also on the frontal-view VTON task using VITON-HD~\cite{lee2022high} and DressCode~\cite{morelli2022dress} datasets.
The results demonstrate that our method outperforms existing methods on both tasks.
% , quantitatively and qualitatively.

In summary, our contributions are outlined below:
 
\begin{itemize}[leftmargin=*]
\item We introduce a novel Multi-View Virtual Try-ON (MV-VTON) task, which aims at generating realistic dressing-up results of the multi-view person by using the given frontal and back clothing.

\item We propose a view-adaptive selection method, where hard-selection and soft-selection are applied to global and local clothing feature extraction, respectively. It ensures that the clothing features are roughly fit to the person's view.

\item We propose joint attention blocks to align the global and local features of selected clothing with the person ones, and fuse them. 

\item We collect a multi-view virtual try-on dataset. Extensive experiments demonstrate that our method outperforms previous approaches quantitatively and qualitatively in both frontal-view and multi-view virtual try-on tasks.
\end{itemize}

\section{Related Work}
\subsection{GAN-Based Virtual Try-On}
Existing methods are aimed at the frontal-view VTON task. To reconstruct realistic results, these methods based on generative adversarial networks (GAN)~\cite{goodfellow2020generative} are typically divided into two steps. 
Firstly, the frontal-view clothing is deformed to align with the target person's pose. Afterward, the warped clothing and target person are fused through a GAN-based generator. In the warping step, some methods~\cite{yang2020towards, ge2021disentangled, wang2018toward} use TPS transformation to deform the frontal-view clothing, and others~\cite{lee2022high, ge2021parser, xie2023gp} predict the global and local optical flow required for clothing deformation. 
However, when the clothing possesses intricate high-frequency details and the person's pose is complex, the effectiveness of clothing deformation is often diminished. Moreover, GAN-based generators generally encounter challenges in convergence and are highly susceptible to mode collapse~\cite{miyato2018spectral}, leading to noticeable artifacts at the junction between warped clothing and the target person in the final results. In addition, previous multi-pose virtual try-on methods~\cite{mp1, mp2, mp3} can change the person's pose, but are also limited by GAN-based generator and insufficient clothing information.

\subsection{Diffusion-Based Virtual Try-On}

Thanks to the rapid advancement of diffusion models, recent works have sought to utilize the generative prior of large-scale pre-trained diffusion models~\cite{ho2020denoising, song2020denoising, rombach2022high, yang2023paint} to tackle frontal-view virtual try-on tasks. TryOnDiffusion~\cite{zhu2023tryondiffusion} introduces two U-Nets to encode target person and frontal-view clothing images respectively, and interacts with the features of the two branches through the cross-attention mechanism. 

LaDI-VTON~\cite{morelli2023ladi} encodes the frontal-view clothing image through textual inversion~\cite{gal2022image, wei2023elite} and serves as the conditional input of backbone. DCI-VTON~\cite{gou2023taming} first conducts an initial deformation of frontal-view clothing by incorporating a pre-trained wrapping network~\cite{ge2021parser}. Subsequently, it attaches the deformed clothing to the target person image and feeds it into the diffusion model. 
While their frontal-view virtual try-on results seem more natural compared to GAN-based methods, they face difficulties in preserving high-frequency details due to the loss of details from the CLIP image encoder~\cite{radford2021learning}. To address this problem, StableVITON~\cite{kim2023stableviton} attempts to introduce an additional encoder~\cite{Zhang_2023_ICCV} to encode the features of frontal-view clothing, and align the obtained clothing features through the zero cross-attention block. However, due to the absence of adequate clothing priors, the generated results often struggle to remain faithful to the original clothing. 
% Additionally, lacking direct interaction between the clothing features makes it easy to lose detailed information. 
Therefore, we introduce joint attention blocks to extract the global and local features of clothing, and employ the view-adaptive selection to choose the clothing features from the two views.

\section{Method}

\subsection{Preliminaries for Diffusion Models} \label{subsec: sd}
Diffusion Models~\cite{ho2020denoising, rombach2022high} have demonstrated strong capabilities in visual generation, which transforms a Gaussian distribution into a target distribution by iterative denoising.
In particular, Stable Diffusion~\cite{rombach2022high} is a widely used generative diffusion model, which consists of a CLIP text encoder $\mathcal{E}_T$, a VAE encoder $\mathcal{E}$ as well as decoder $\mathcal{D}$, and a time-conditional denoising model $\epsilon_{\theta}$. The text encoder $\mathcal{E}_T$ encodes the input text prompt $y$ as conditional input. The VAE encoder $\mathcal{E}$ compresses the input image $I$ into latent space to get the latent variable $z_0=\mathcal{E}(I)$. In contrast, the VAE decoder $\mathcal{D}$ decodes the output of backbone from latent space to pixel space. Through the VAE encoder $\mathcal{E}$, at an arbitrary time step t, the forward process is performed:
\begin{equation}
\alpha : =  {\textstyle \prod_{s=1}^{t}(1-\beta _s)} ~, \indent
z_t = \sqrt{\alpha_t}z_0 + \sqrt{1-\alpha_t}\epsilon, 
    \label{con:ldm forward}
\end{equation}
where $\epsilon \sim \mathcal{N}(0,1)$ is the random Gaussian noise and $\beta$ is a predefined variance schedule.
The training objective is to acquire a noise prediction network that minimizes the disparity between the predicted noise and the noise added to ground truth. The loss function can be defined as,
\begin{equation}
\mathcal{L}_{LDM} = \mathbb{E}_{\mathcal{E}(I),y,\epsilon \sim \mathcal{N}(0,1),t}[\left \| \epsilon - \epsilon_{\theta } (z_t,t,\mathcal{E}_T(y)) \right \|_2^2 ],
\label{con:ldm_loss}
\end{equation}
where $z_t$ represents the encoded image $\mathcal{E}(I)$ with random Gaussian noise $\epsilon \sim \mathcal{N}(0,1)$ added.

In our work, we use an exemplar-based inpainting model~\cite{yang2023paint} as a backbone, which employs an image $c$ rather than texts as the prompt and then encode $c$ by the image encoder $\mathcal{E}_I$ of CLIP. Thus, the loss function in Eq.~(\ref{con:ldm_loss}) can be modified as,
\begin{equation}
\mathcal{L}_{LDM} = \mathbb{E}_{\mathcal{E}(I),c,\epsilon \sim \mathcal{N}(0,1),t}[\left \| \epsilon - \epsilon_{\theta } (z_t,t,\mathcal{E}_I(c)) \right \|_2^2 ].
\label{con:pbe_loss}
\end{equation}
% where $c$ is the image prompt.

\subsection{Method Overview} 

\begin{figure}[t!]
    \includegraphics[width=0.479\textwidth]{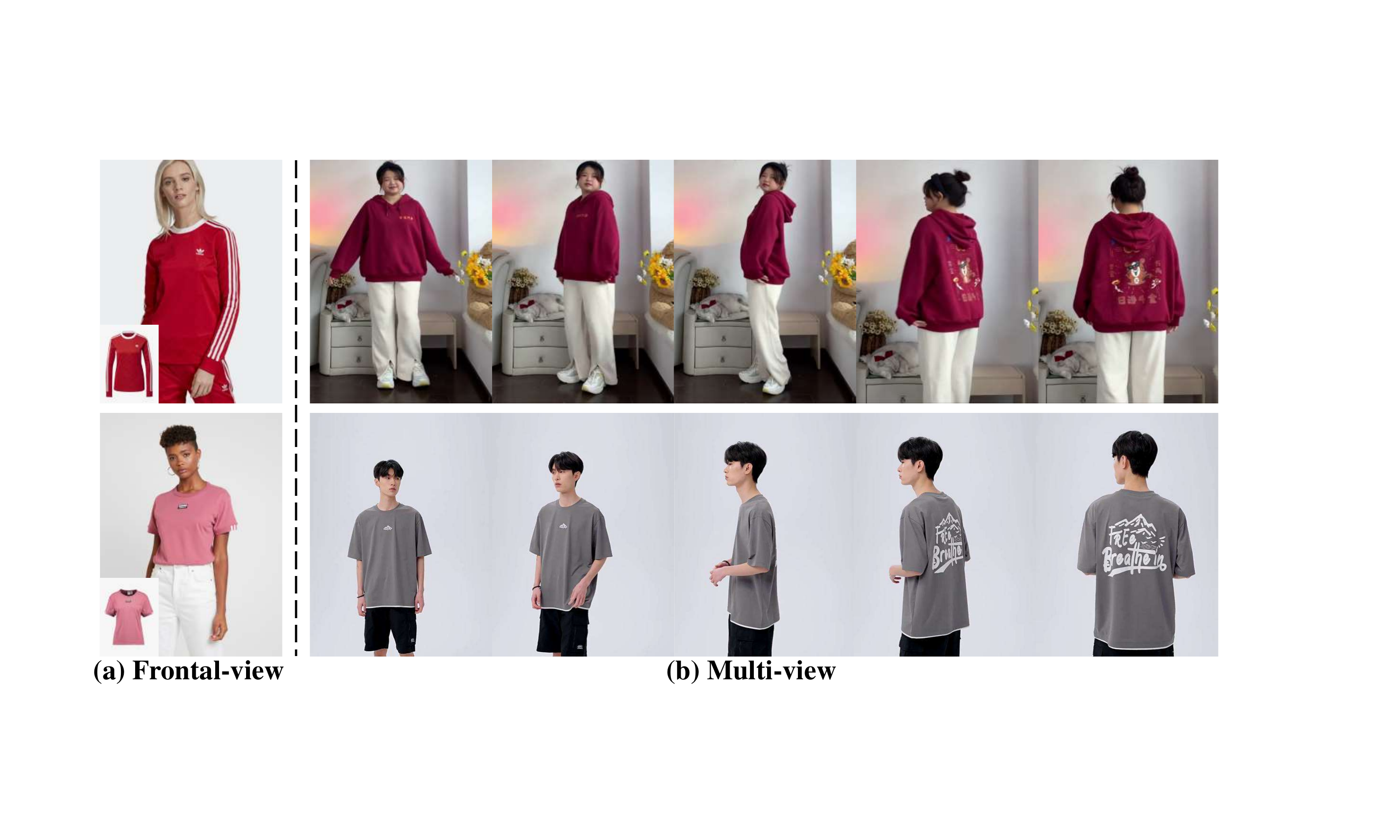}
    \caption{Comparison between previous datasets and our proposed MVG dataset. (a) is the dataset used by the previous work, which only have clothing and person in the frontal-view. In contrast, our dataset (b) offers images from five different views. 
    }
    \label{fig:dataset_show}
\end{figure}

While existing virtual try-on methods are designed solely for frontal-view scenarios, we present a novel approach to handle both frontal-view and multi-view virtual try-on tasks, along with a multi-view virtual try-on dataset MVG comprising try-on images captured from five different views. Examples of it are shown in Figure~\ref{fig:dataset_show}(b). Formally, given a person image $x$
% $x \in R^{H\times W\times 3}$ 
in an arbitrary view, along with a frontal view clothing $c_{f}$ and a back view clothing $c_{b}$, our goal is to generate the result of the person wearing the clothing in its view. Considering the substantial differences between the front and back of most clothing, another challenge is to make informed decisions regarding the two provided clothing images based on the target person's pose, ensuring a natural try-on result across multiple views.

\begin{figure*}[t]
\centering
    \includegraphics[width=1\textwidth]{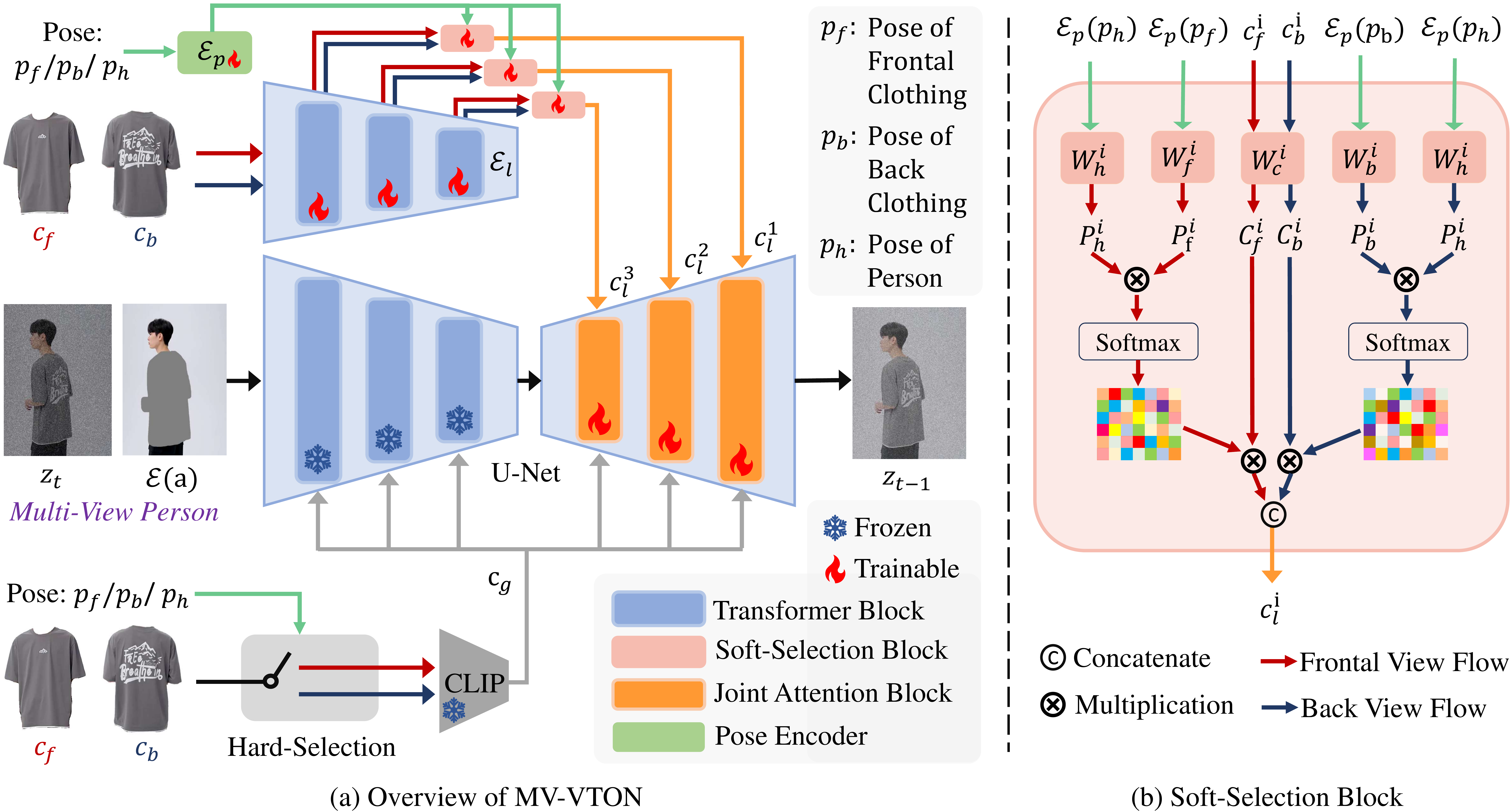}
    \caption{(a) Overview of MV-VTON. It encodes frontal and back view clothing into global features using the CLIP image encoder and extracts multi-scale local features through an additional encoder~$\mathcal{E}_l$. Both features act as conditional inputs for the decoder of backbone. Besides, both features are selectively extracted through view-adaptive selection mechanism. (b) Soft-selection modulates the clothing features on frontal and back views, respectively, based on the similarity between the clothing's pose and the person's pose. Then the features from both views are concatenated in the channel dimension.}
    
    \label{fig:framework}
\end{figure*}

In this work, we use an image inpainting diffusion model~\cite{yang2023paint} as our backbone. 
Denote by $M$ the inpainting mask, and denote by $a$ the masked person image $x$.
The model concatenates $z_t$ ($z_0=\mathcal{E}(x)$), the encoded clothing-agnostic image $\mathcal{E}(a)$, and the resized clothing-agnostic mask $m$ in the channel dimension, and feeds them into the backbone as spatial input. Besides, we use an existing method to pre-warp the clothing and paste it on $a$. 
% $\mathcal{E}$ represents VAE encoder~\cite{vae} in the latent diffusion model. 
While utilizing CLIP image encoder to encode clothing as the global condition of the diffusion model, we also introduce an additional encoder~\cite{Zhang_2023_ICCV} to encode clothing to provide more refined local conditions. Since both the frontal and back view clothing need to be encoded, directly sending both into the backbone as conditions may result in confusion of clothing features. To alleviate this problem, we propose a view-adaptive selection mechanism. Based on the similarity between the poses of the person and two clothes, it conducts hard-selection when extracting global features and soft-selection when extracting local features. To preserve semantic information in clothing and enhance high-frequency details in global features using local ones, we introduce joint attention blocks. They first independently align global and local features to the person ones and then selectively fuse them. Figure~\ref{fig:framework}(a) depicts an overview of our proposed method.

\subsection{View-Adaptive Selection}

For multi-view virtual try-on task, given the substantial differences between the frontal and back views, as illustrated in Figure~\ref{fig:dataset_show}(b), it's imperative to extract and assign the features of frontal and back view clothing for the person tendentiously.
Actually, based on the pose of the target person, we can determine which view of clothing should be given more attention during the try-on process. 
For example, if the target pose resembles the pose in the fourth column of Figure~\ref{fig:dataset_show}(b), it's evident that we should rely more on the characteristics of the back view clothing to generate the try-on result. 
Specifically, we propose a view-adaptive selection mechanism to achieve this purpose, including hard- and soft-selection.

\noindent \textbf{Hard-Selection for Global Clothing Features.} 
We deploy a CLIP image encoder to extract global features of clothing.
During this process, we perform hard-selection on the frontal and back view clothing based on the similarity between the garments' pose and the person's pose. It means that we only select one piece of clothing that is closest to the person's pose as the input of the image encoder, since it is enough to cover global semantic information. When generating pre-warped clothing for $\mathcal{E}(a)$, the selection is also performed. Implementation details of hard-selection can be found in the supplementary material.

\noindent \textbf{Soft-Selection for Local Clothing Features.} 
We utilize an additional encoder $\mathcal{E}_l$ to extract the multi-scale local features of frontal and back view clothing, which in the $i$-th scale are denoted as $c_f^i$ and $c_b^i$, respectively.
When reconstructing the try-on results, it may be insufficient to rely solely on the clothing from either frontal or back view under certain specific scenes, such as the third column shown in Figure~\ref{fig:dataset_show}(b). 
In these cases, it may be necessary to incorporate clothing features from both views. 
However, simply combining the two may lead to confusion of features.
Instead, we introduce soft-selection block to modulate their features, respectively, as shown in Figure~\ref{fig:framework}(b).
First, the person's pose $p_h$, frontal-view clothing's pose $p_f$, and back view clothing's pose $p_b$ are encoded by the pose encoder $\mathcal{E}_p$ to obtain their respective features $\mathcal{E}_p(p_h)$, $\mathcal{E}_p(p_f)$, and $\mathcal{E}_p(p_b)$. Details of the pose encoder can be found in the supplementary material.
When processing frontal-view clothing, in $i$-th soft-selection block, we map $\mathcal{E}_p(p_h)$ and $\mathcal{E}_p(p_f)$ to $P^i_h$ and $P^i_f$ through a linear layer with weights $W_h^i$ and $W_f^i$, respectively. We also map $c_f^i$ to $C^i_f$ through a linear layer with weights $W_c^i$.
Then, we calculate the similarity between the person's pose and frontal-view clothing's pose to get the selection weights of frontal-view clothing, \ie,
\begin{equation}
    weights = softmax(\frac{P_h^i(P_f^i)^T}{\sqrt{d}}),
    \label{con:soft}
\end{equation}
where $weights$ represents the selection weights of frontal-view clothing, and $d$ represents the dimension of these matrices. Assuming that the person's pose is biased towards the front, as depicted in the second column of Figure~\ref{fig:dataset_show}(b), the similarity between the person's pose and the front view clothing's pose will be higher. Consequently, the corresponding clothing features will be enhanced by $weights$, and vice versa.
The features of back view clothing $c_b^i$ undergo similar processing.
Finally, the two selected clothing features are concatenated along the channel dimension as the local condition $c_l^i$ of backbone.

\begin{figure}[t!]
    \includegraphics[scale=0.9]{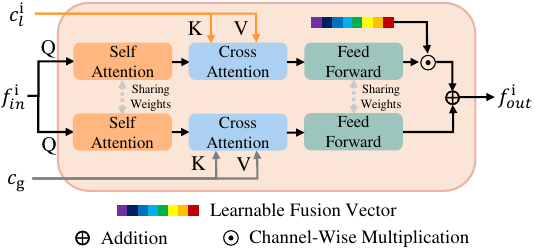}
    \caption{Overview of the proposed joint attention blocks.}
    \label{fig:jab}
\end{figure}

\subsection{Joint Attention Blocks}
Global clothing features $c_g$ provide identical conditions for blocks at each scale of U-Net, and multi-scale local clothing features $c_l$ allow for reconstructing more accurate details.
We present joint attention blocks to align $c_g$ and $c_l$ with the current person features, as shown in Figure~\ref{fig:jab}.
To retain most of the semantic information in global features $c_g$, we use local features $c_l$ to refine some lost and erroneous detailed texture information in $c_g$ by selective fusion.

Specifically, in the $i$-th joint attention block, we first calculate self-attention for the current features $f_{in}^i$. Then, we deploy a double cross-attention. The queries (Q) come from $f_{in}^i$ and global features $c_g$ serve as one set of keys (K) and values (V), while local features $c_l^i$ serve as another set of keys (K) and values (V). After aligning to the person's pose through cross-attention, the clothing features $c_g$ and $c_l^i$ are selectively fused in channel-wise dimension, \ie,
\begin{equation}
\begin{aligned}
f_{out}^i = ~ & softmax(\frac{Q^i_g(K^i_g)^T}{\sqrt{d} } )V^i_g ~ + 
\\ & \lambda \odot softmax(\frac{Q^i_l(K^i_l)^T}{\sqrt{d} } )V^i_l ~,
\label{con:jab}
\end{aligned}
\end{equation}
where $Q^i_g, K^i_g, V^i_g$ represent the Q, K, V of global branch, $Q^i_l, K^i_l, V^i_l$ represent the Q, K, V of local branch, $\lambda$ is the learnable fusion vector, $\odot$ represents channel-wise multiplication, and $f_{out}^i$ represents the clothing features after selective fusion. By engaging and fusing the global and local clothing features, we can enhance the retention of high-frequency garment details, \eg, texts and patterns.
% , ensuring their fidelity within the garment.

\subsection{Training Objectives}
As stated in preliminaries, diffusion models learn to generate images from random Gaussian noise. However, the training objective in Eq.~(\ref{con:pbe_loss}) is performed in latent space, and does not explicitly constrain the generated results in visible image space, resulting in slight differences in color from the ground truth. To alleviate the problem, we additionally employ $\ell_1$ loss $\mathcal{L}_1$ and perceptual loss~\cite{perceptual} $\mathcal{L}_{perc}$. The $\mathcal{L}_1$ loss is calculated by
\begin{equation}
\mathcal{L}_1 = \left \| \hat{x} - x  \right \| _1, 
    \label{con:l1}
\end{equation}
where $\hat{x}$ is the reconstructed image using Eq.~(\ref{con:ldm forward}). The perceptual loss is calculated as,
\begin{equation}
\mathcal{L}_{perc} = \sum_{k=1}^{5}\left \| \phi _k(\hat{x} ) - \phi _k(x ) \right \|  _1, 
    \label{con:vgg}
\end{equation}
where $\phi _k$ represents the $k$-th layer of VGG~\cite{vgg}. Totally, the overall training objective can be written as,
\begin{equation}
\mathcal{L} =  \mathcal{L}_{LDM} + \lambda _1 \mathcal{L}_1 + \lambda _{perc} \mathcal{L}_{perc}~, 
    \label{con:total loss}
\end{equation}
where $\lambda _1$ and $\lambda _{perc}$ are the balancing weights.

\section{Experiments}
\subsection{Experiments Settings}
\noindent \textbf{Datasets:} For the proposed multi-view virtual try-on task, we collect MVG dataset containing 1,009 samples. Each sample contains five images of the same person wearing the same garment from five different views, for a total of 5,045 images, as shown in Figure~\ref{fig:dataset_show}(b). The image resolution is about 1K. We'll explain how the datasets are collected and how they're used for MV-VTON in the supplementary material.
The proposed method can also be applied to frontal-view virtual try-on task. Our frontal-view experiments are carried out on VITON-HD~\cite{lee2022high} and DressCode~\cite{morelli2022dress} datasets. They contain more than 10,000 frontal-view person and upper-body clothing image pairs. We follow previous work for the use of them.

\noindent \textbf{Evaluation Metrics.} Following previous works~\cite{kim2023stableviton, morelli2023ladi}, we use four metrics to evaluate the performance of our method: Structural Similarity (SSIM)~\cite{ssim}, Learned Perceptual Image Patch Similarity (LPIPS)~\cite{lpips}, Frechet Inception Distance (FID)~\cite{fid} and Kernel Inception Distance (KID)~\cite{kid}.
Specifically, for paired test setting, which means directly using the paired data in the dataset, we utilize the above four metrics for evaluation. For unpaired test setting, which means that the given garment is different from the garment originally worn by target person, we use FID and KID for evaluation, and in order to distinguish them from the paired setting, we named them ${\rm FID_u}$ and ${\rm KID_u}$ respectively.

\begin{table*}
\centering
\resizebox{\textwidth}{!}{
\begin{tabular}{l|l|cccc|cccc|cccc}
\toprule
\multirow{2}{*}{Methods} & \multirow{2}{*}{Reference} & \multicolumn{4}{c|}{MVG} & \multicolumn{4}{c|}{VITON-HD} & \multicolumn{4}{c}{DressCode - Upper Body}\\
 & &  LPIPS↓ & SSIM↑ & FID↓ & KID↓ & LPIPS↓ & SSIM↑ & FID↓ & KID↓ & LPIPS↓ & SSIM↑ & FID↓ & KID↓\\
\midrule
Paint by Example & CVPR23 & 0.120 & 0.880 & 54.38 & 14.95 & 0.150 & 0.843 & 13.78 & 4.48 & 0.078 & 0.899 & 15.21 & 4.51\\
PF-AFN & CVPR21 & 0.139 & 0.873 & 49.47 & 12.81 & 0.141 & 0.855 & 7.76 & 4.19 & 0.091 & 0.902 & 13.11 & 6.29\\
GP-VTON & CVPR23 & - & - & - & - & 0.085 & 0.889 & 6.25 & 0.77 & 0.236 & 0.781 & 19.37 & 8.07\\
LaDI-VTON & MM23 & 0.069 & 0.921 & 29.14 & 4.39 & 0.094 & 0.872 & 7.08 & 1.49 & 0.063 & 0.922 & 11.85 & 3.20 \\
DCI-VTON & MM23 & 0.062 & 0.929 & 25.71 & 0.95 & 0.074 & 0.893 & 5.52 & 0.57 & 0.043 & 0.937 & 11.87 & 1.91\\
StableVITON & CVPR24 & 0.063 & 0.929 & 23.52 & 0.46 & 0.073 & 0.888 & 6.15 & 1.34 & \textbf{0.040} & 0.937 & 10.18 & 1.70\\
IDM-VTON & ECCV24 & 0.095 & 0.896 & 34.66 & 5.33 & 0.135 & 0.826 & 14.36 & 8.63 & 0.066 & 0.912 & 13.88 & 5.39\\
Ours & - & \textbf{0.050} & \textbf{0.936} & \textbf{22.18} & \textbf{0.35} & \textbf{0.069} & \textbf{0.897} & \textbf{5.43} & \textbf{0.49} & \textbf{0.040} & \textbf{0.941} & \textbf{8.26} & \textbf{1.39} \\
\bottomrule
\end{tabular} 
}
\caption{Quantitative comparison with previous work on paired setting. For multi-view virtual try-on task, we show results on our proposed MVG dataset. For frontal-view virtual try-on task, we show results on VITON-HD dataset~\cite{lee2022high} and DressCode dataset~\cite{morelli2022dress}. The best results have been bolded. Note that all previous works have been finetuned on our proposed MVG dataset when comparing on multi-view virtual try-on task.}
\label{table:cmp results}
\end{table*}

\begin{figure*}[t!]
\centering
    \includegraphics[width=0.95\textwidth]{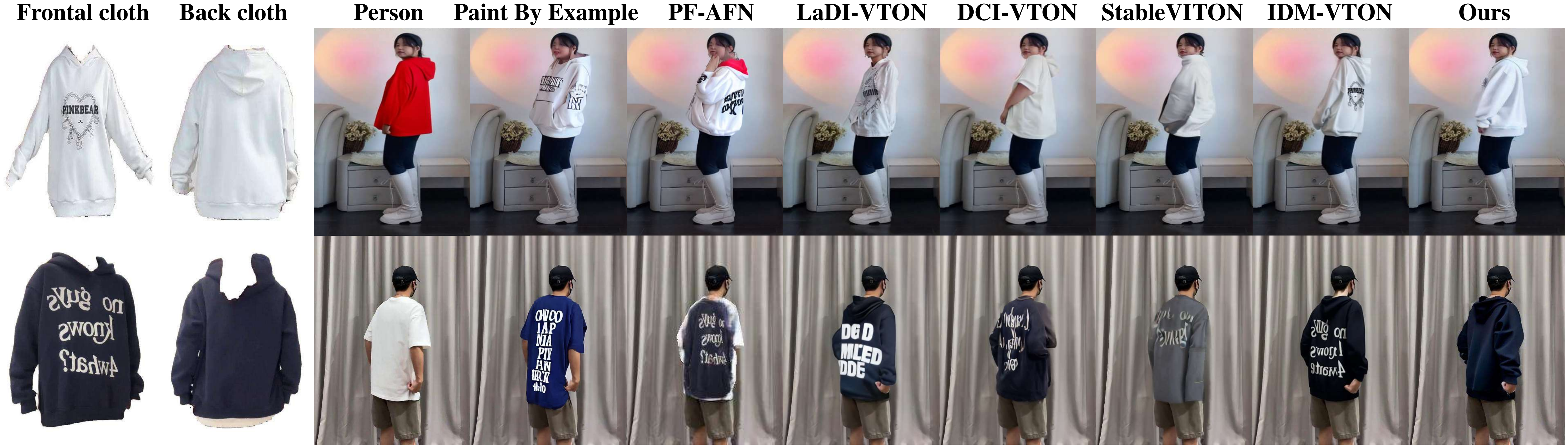}
    \caption{Qualitative comparisons on multi-view virtual try-on task with MVG dataset.}
    \label{fig:cmp_mv}
    
\centering
    \includegraphics[width=0.95\textwidth]{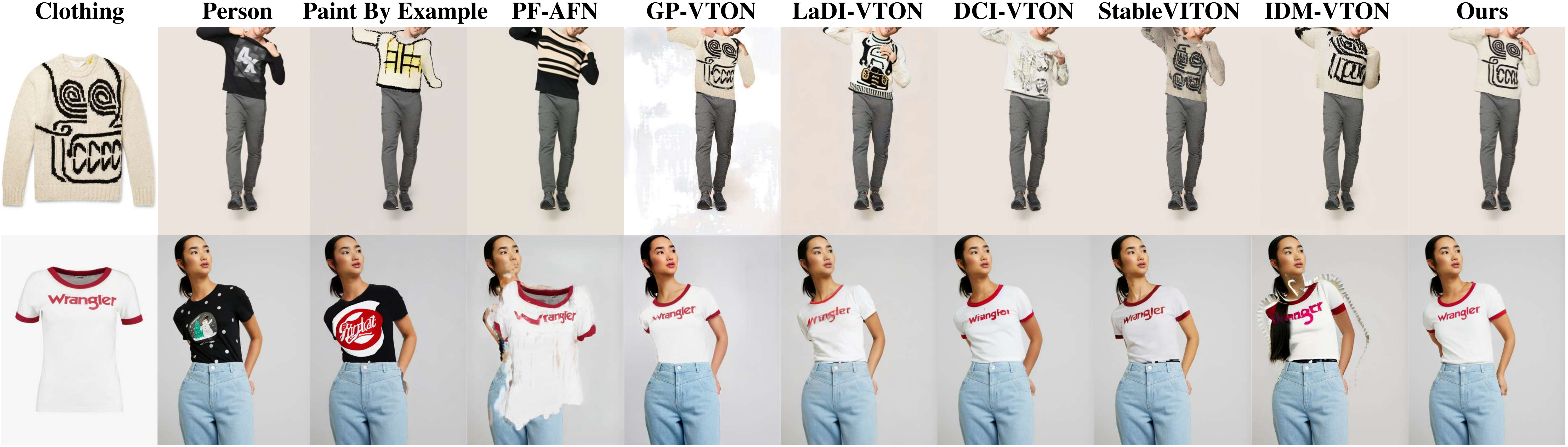}
    \caption{Qualitative comparisons on frontal-view virtual try-on task with VITON-HD and DressCode datasets.}
    \label{fig:cmp_viton}
\end{figure*}

\noindent \textbf{Implementation Details.}
We use Paint by Example~\cite{yang2023paint} as the backbone of our method and copy the weights of its encoder to initialize $\mathcal{E}_l$. The hyper-parameter $\lambda_1$ is set to 1e-1, and $\lambda_{perc}$ is set to 1e-4. We train our model on 2 NVIDIA Tesla A100 GPUs for 40 epochs with a batch size of 4 and a learning rate of 1e-5. We use AdamW~\cite{adamW} optimizer with $\beta_1=0.9$, $\beta_2=0.999$.

\noindent \textbf{Comparison Settings.} 
We compare our method with Paint By Example~\cite{yang2023paint}, PF-AFN~\cite{ge2021parser}, GP-VTON~\cite{xie2023gp}, LaDI-VTON~\cite{morelli2023ladi}, DCI-VTON~\cite{gou2023taming}, StableVITON~\cite{kim2023stableviton} and IDM-VTON~\cite{IDM-VTON} on both frontal-view and multi-view virtual try-on tasks.
For multi-view virtual try-on, we compare these methods on the proposed MVG dataset. For the sake of fairness, we fine-tune the previous methods on the MVG dataset according to its original training settings. Since previous methods can only input a single clothing image, we input frontal and back view clothing respectively and select the best result. For frontal-view virtual try-on, we compare these methods on VITON-HD and DressCode datasets. Following previous works' settings, the proposed MV-VTON only inputs one frontal-view garment during training and inference.

\subsection{Quantitative Evaluation}
Table~\ref{table:cmp results} reports the quantitative results on the paired setting, and Table~\ref{table:unpaired} shows the unpaired setting’s results. On the multi-view virtual try-on task, as can be seen, thanks to the view-adaptive selection mechanism, our method can reasonably select clothing features according to the person's pose, so it is better than existing methods in various metrics, especially on LPIPS and SSIM. Furthermore, owing to joint attention blocks, our approach excels in preserving high-frequency details of the original garments across both frontal-view and multi-view virtual try-on scenarios, thus achieving superior performance in these metrics.

\subsection{Qualitative Evaluation}
\noindent \textbf{Multi-View Virtual Try-On.} As shown in Figure~\ref{fig:cmp_mv}, MV-VTON generates more realistic multi-view results compared to the previous five methods. Specifically, in the first row, due to the lack of adaptive selection of clothes, previous methods have difficulty in generating hoods of the original cloth. Moreover, in the second row, previous methods often struggle to maintain fidelity to the original garments. In contrast, our method effectively addresses the aforementioned problems and generates high-fidelity results. We provide more results of multi-view virtual try-on in the supplementary materials.

\begin{table}[t!]
\centering
\begin{tabular}{c|cc|cc}
\toprule
\multirow{2}{*}{Method} & \multicolumn{2}{c|}{MVG} & \multicolumn{2}{c}{VITON-HD} \\
 & ${\rm FID_u}$↓ & ${\rm KID_u}$↓ & ${\rm FID_u}$↓ & ${\rm KID_u}$↓ \\
\midrule
Paint by Example & 43.79 & 5.92 & 17.27 & 4.56\\
PF-AFN & 47.38 & 7.04 & 21.18 & 6.57 \\
GP-VTON & - & - & 9.11 & 1.21 \\
LaDI-VTON & 36.61 & 3.39 & 9.55 & 1.83 \\
DCI-VTON & 36.03 & 3.79 & 8.93 & 1.07 \\
StableVITON & 35.85 & 4.22 & 9.86 & 1.09 \\
IDM-VTON & 40.73 & 5.74 & 18.27 & 10.43 \\
Ours & \textbf{33.44} & \textbf{2.69} & \textbf{8.67} & \textbf{0.78} \\
\bottomrule
\end{tabular}
\caption{Unpaired setting's quantitative results on our MVG dataset and VITON-HD dataset. The best results have been bolded. }
\label{table:unpaired}
\end{table}

\begin{table}[t!]
\centering
\resizebox{0.48\textwidth}{!}{
\begin{tabular}{cc|cccccc}
\toprule
Hard & Soft & LPIPS↓ & SSIM↑ & FID↓ & KID↓ & ${\rm FID_u}$↓ & ${\rm KID_u}$↓ \\
\midrule
$\times $ & $\times $ & 0.068 & 0.925 & 25.13 & 0.77 & 35.28 & 3.24 \\
$\times $ & $\surd $ & 0.064 & 0.928 & 24.58 & 0.62 & 34.67 & 3.05 \\
$\surd $ & $\times $ & 0.052 & 0.934 & \textbf{22.18} & 0.43 & 33.47 & 2.74 \\
$\surd $ & $\surd $ & \textbf{0.050} & \textbf{0.936} & \textbf{22.18} & \textbf{0.35} & \textbf{33.44} & \textbf{2.69} \\
\bottomrule
\end{tabular}
}
\caption{Ablation study of our proposed view-adaptive selection mechanism on MVG dataset.}
\label{table:ab pose}
\end{table}

\begin{table}[t!]
\renewcommand{\arraystretch}{1.18}
\centering
\resizebox{0.48\textwidth}{!}{
\begin{tabular}{c|cc|cccccc}
\toprule
 & Global & Local & LPIPS↓ & SSIM↑ & FID↓ & KID↓ & ${\rm FID_u}$↓ & ${\rm KID_u}$↓\\
 \midrule
\multirow{3}{*}{\rotatebox[origin=c]{90}{\scriptsize MVG}}
 & $\surd $ & $\times $ & 0.062 & 0.929 & 25.71 & 0.95 & 36.01 & 3.78 \\
 & $\times $ & $\surd $ & 0.058 & 0.931 & 26.16 & 1.21 & 36.29 & 3.91\\
 & $\surd $ & $\surd $ & \textbf{0.050} & \textbf{0.936} & \textbf{22.18} & \textbf{0.35} & \textbf{33.44} & \textbf{2.69} \\
\midrule
\multirow{3}{*}{\rotatebox[origin=c]{90}{\scriptsize VITON-HD}}
 & $\surd $ & $\times $ & 0.074 & 0.893 & 5.52 & 0.57 & 8.93 & 1.07 \\
 & $\times $ & $\surd $ & 0.070 & 0.896 & 5.76 & 0.81 & 9.15 & 1.09 \\
 & $\surd $ & $\surd $ & \textbf{0.069} & \textbf{0.897} & \textbf{5.43} & \textbf{0.49} & \textbf{8.67} & \textbf{0.78} \\
\bottomrule
\end{tabular}
}
\caption{Ablation study of joint attention blocks on MVG and VITON-HD datasets.}
\label{table:ab viton}
\end{table}

\noindent \textbf{Frontal-View Virtual Try-On.} As shown in Figure~\ref{fig:cmp_viton}, our method also demonstrates superior performance over existing methods on frontal-view virtual try-on task, particularly in retaining clothing details. Specifically, our method not only faithfully generates complex patterns (in the first row), but also better preserves the literal 'Wrangler' in the clothing (in the second row). We provide more qualitative comparisons in the supplementary materials, as well as dressing results under complex human pose conditions.

\subsection{Ablation Studies}
\begin{figure}[t!]
\centering
    \includegraphics[width=0.475\textwidth]{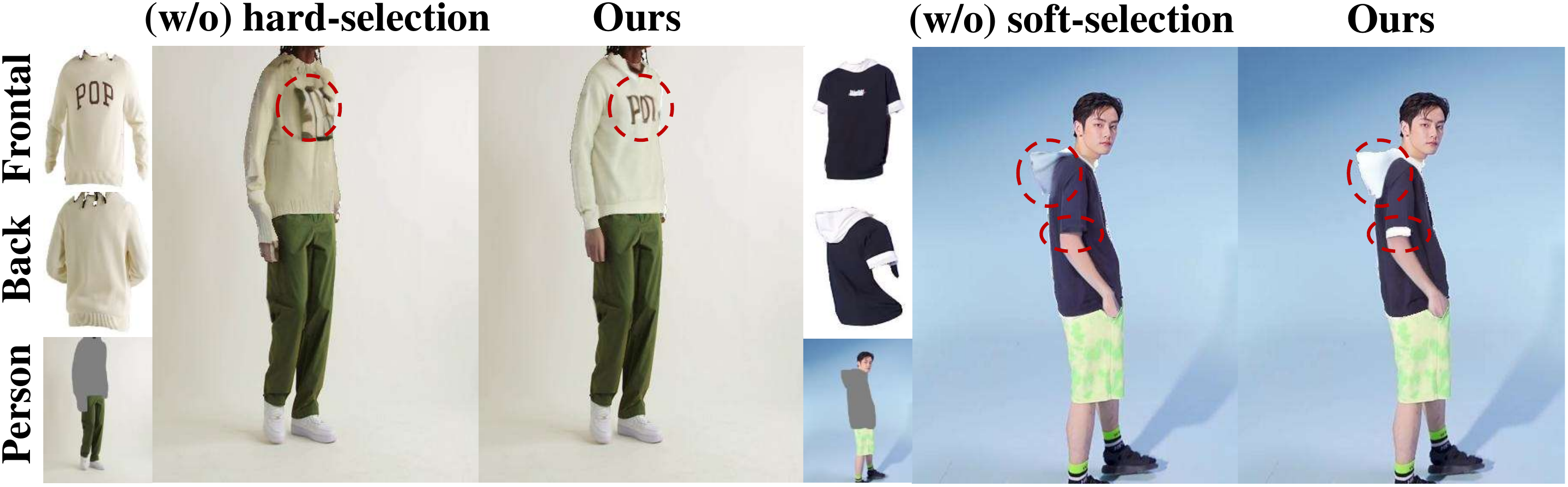}
    \caption{Visualization of view-adaptive selection's effect.}
    \label{fig:ab pose}
\end{figure}

\noindent \textbf{Effect of View-Adaptive Selection.} We investigate the effect of view-adaptive selection on the multi-view virtual try-on task. Specifically, no hard-selection represents that we directly concatenate two garments' features encoded by CLIP, and no soft-selection means that two clothing features are concatenated without passing soft-selection blocks. Comparison results are shown in Table~\ref{table:ab pose} and Figure~\ref{fig:ab pose}. As can be seen, the performance is greatly reduced without hard-selection and soft-selection. No hard-selection will confuse two view's cloth features, as shown by the blurriness of the 'POP' text in Figure~\ref{fig:ab pose}. In addition, no soft-selection causes the model to lose some cloth information when processing the side view situation, such as the missing white hood and cuffs in Figure~\ref{fig:ab pose}.

\begin{figure}[t!]
\centering
    \includegraphics[width=0.475\textwidth]{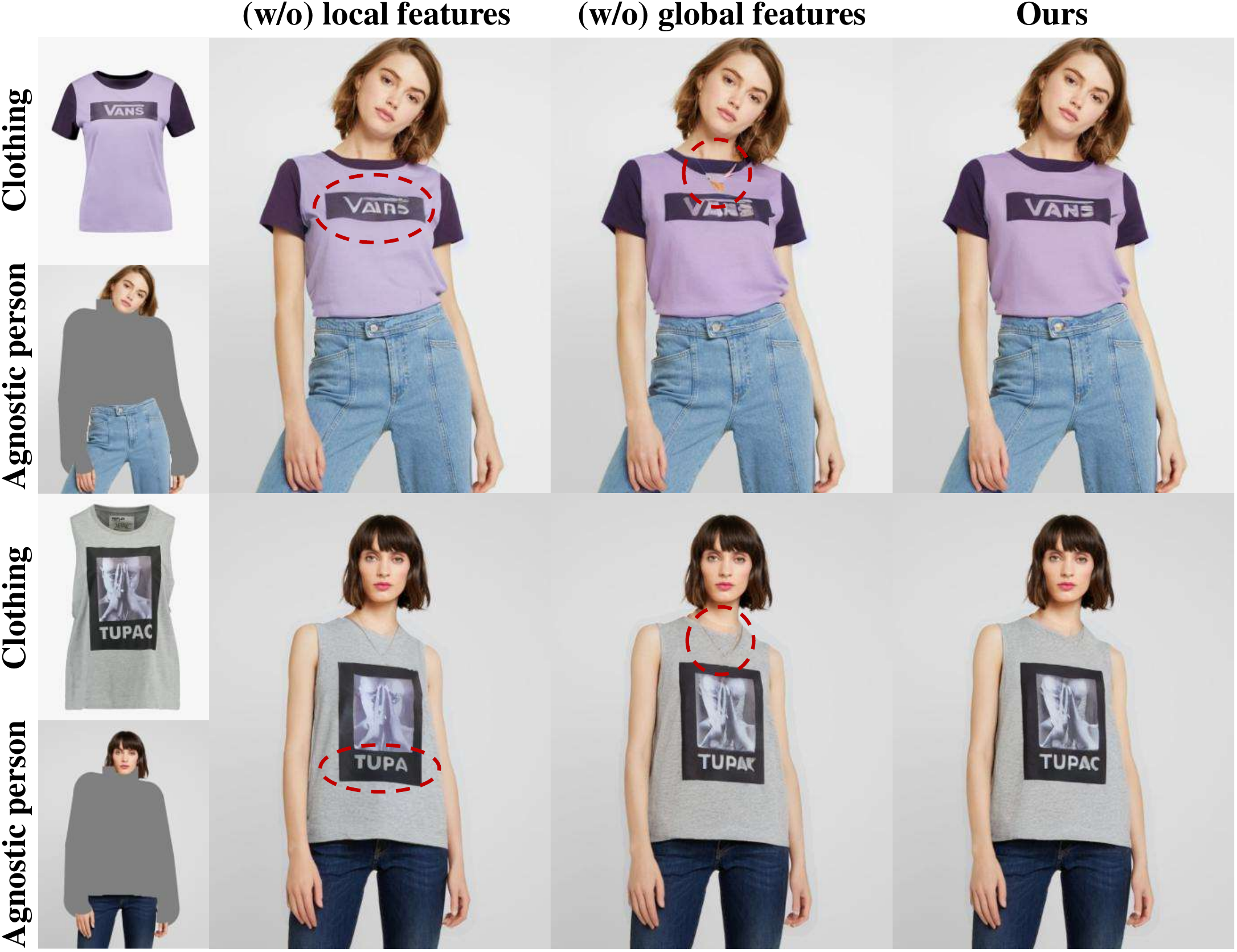}
    \caption{Visualization of joint attention blocks' effect.}
    \label{fig:ab jab}
\end{figure}

% fig
\noindent \textbf{Effect of Joint Attention Blocks.} In order to demonstrate the effectiveness of fusing global and local features through joint attention blocks, we discard the global feature extraction branch and the local feature extraction branch respectively. Results are shown in Table~\ref{table:ab viton} and Figure~\ref{fig:ab jab}. As can be seen, relying solely on global features may lead to loss of details, such as the distorted text 'VANS' in the first row and the missing letter 'C' in the second row. Moreover, if only local features are provided, the results may also have unfaithful textures, such as artifacts on the person's chest. Compared to them, we fuse global and local features through joint attention blocks, which can refine details in garments while preserving semantic information.

\section{Conclusion}
We introduce a novel and practical Multi-View Virtual Try-ON (MV-VTON) task, which aims at using the frontal and back clothing to reconstruct the dressing results of a person from multiple views.
To achieve the task, we propose a diffusion-based method.
Specifically, the view-adaptive selection mechanism exacts more reasonable clothing features based on the similarity between the poses of a person and two clothes. The joint attention block aligns the global and local features of the selected clothing to the target person, and fuse them.
In addition, we collect a multi-view garment dataset for this task.
Extensive experiments demonstrate that the proposed method achieves state-of-the-art performance both on frontal-view and multi-view virtual try-on tasks, compared with existing methods.

\section{Acknowledgments}

This work was supported by the National Key R\&D Program of China (2022YFA1004100).

\appendix
\renewcommand\thesection{\Alph{section}}
\renewcommand{\thesection}{\Alph{section}}
\renewcommand{\thetable}{\Alph{table}}
\renewcommand{\thefigure}{\Alph{figure}}
\renewcommand{\theequation}{\Alph{equation}}

%% 重新计数
\setcounter{section}{0}
\setcounter{figure}{0}

% {\huge \textbf{Supplementary Materials}}

% \section*{\centering{\Large Supplementary Materials\\[30pt]}}
\vspace{2mm}
\section*{Appendix}
\vspace{2mm}

\section{IMPLEMENTATION DETAILS}
\begin{figure}[t!]
\includegraphics[width=0.479\textwidth]{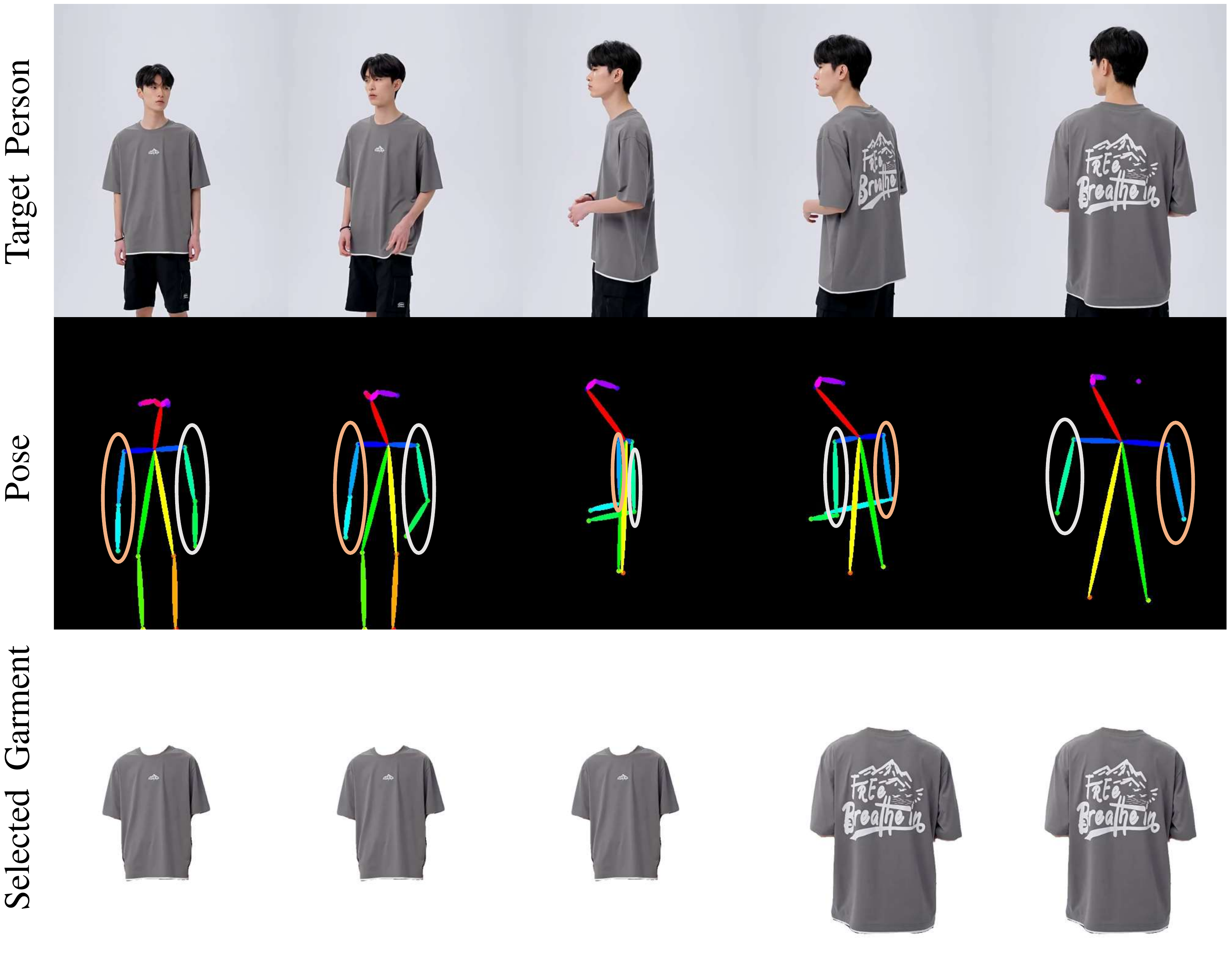}
    \vspace{-6mm}
    \caption{Visualization of the person and corresponding poses. We select one of garments based on the relative positions of left and right arms in the skeleton image when performing hard-selection on the multi-view virtual try-on task.}
    % \vspace{-2mm}
    \label{fig:hard}
\end{figure}

\noindent\textbf{Hard-Selection.}
In this section, we present more details about the proposed hard-selection for global clothing features. Specifically, in multi-view virtual try-on task, we use OpenPose~\cite{cao2017realtime, simon2017hand, wei2016cpm} to extract the skeleton images of target person, frontal clothing and back clothing as pose information $p_h$, $p_f$, and $p_b$, respectively. After that, we decide whether to use frontal-view clothing or back-view clothing based on the relative positions of the target person's left arm and right arm in the skeleton images. As shown in Figure \ref{fig:hard}, if the right arm appears positioned to the left of the left arm in the skeleton image (columns one to three in Figure \ref{fig:hard}), frontal-view clothing is chosen; otherwise, back-view clothing is preferred (columns four to five in Figure \ref{fig:hard}). In addition, following previous works~\cite{gou2023taming, xie2023gp,morelli2023ladi}, we adopt an additional warping network~\cite{ge2021parser, kim2023stableviton} to obtain the pre-warped cloth.

\noindent\textbf{Pose Encoder.}
The pose encoder is used to extract features of skeleton images. It is a tiny network that contains three blocks, followed by the layer normalization~\cite{ba2016layer}. Each block comprises one convolution layer, one GELU~\cite{hendrycks2016gaussian} activation layer, and a down-sampling operation. We utilize the acquired pose embeddings as input for the proposed soft-selection block.

\begin{figure}[t!]
\includegraphics[width=0.479\textwidth]{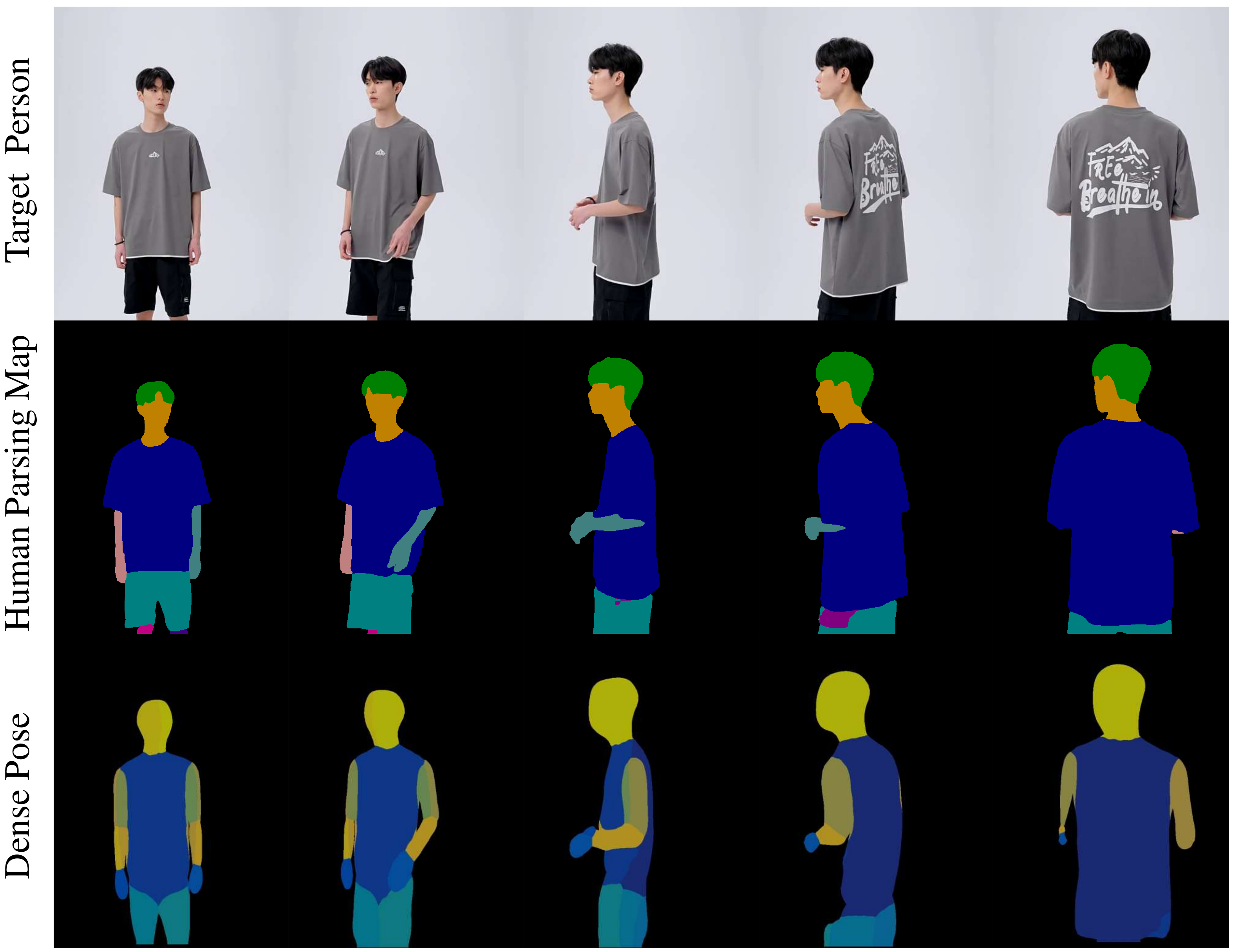}
    \vspace{-6mm}
    \caption{Examples of human parsing maps and dense pose in our dataset. The parsing maps can be used to synthesize cloth-agnostic person images.
    }
    % \vspace{-2mm}
    \label{fig:full_dataset}
\end{figure}

\noindent\textbf{MVG Dataset.}
To construct dataset for multi-view virtual try-on (MV-VTON), we first collect a large number of videos from YOOX NET-A-PORTER\footnote{\url{https://net-a-porter.com}}, Taobao\footnote{\url{https://taobao.com}} and TikTok\footnote{\url{https://douyin.com}} websites, then filter out 1009 videos where the person in the video turns at least 180 degrees while wearing clothes. 
Afterwards, we divide each video into frames and handpick 5 frames to constitute a sample within our MVG dataset. Across these 5 frames, the person is captured from various angles, approximately spanning 0 (\ie, frontal view), 45, 90, 135, and 180 (\ie, back view) degrees, as shown in the first row of Figure~\ref{fig:full_dataset}. In addition, following the previous DressCode dataset~\cite{morelli2022dress}, we employ SCHP model~\cite{li2020self} to extract the corresponding human parsing maps and utilize DensePose~\cite{guler2018densepose} to obtain the person's dense labels, as shown in the second and third row of Figure~\ref{fig:full_dataset}. Human parsing maps can be utilized to generate cloth-agnostic person images, which are necessary for training and inference processes.

% \section{MORE QUANTITATIVE RESULTS}
% \subsection{Results on DressCode}

\section{MORE QUALITATIVE RESULTS}
\begin{figure*}[t!]
\centering
    \includegraphics[width=\textwidth]{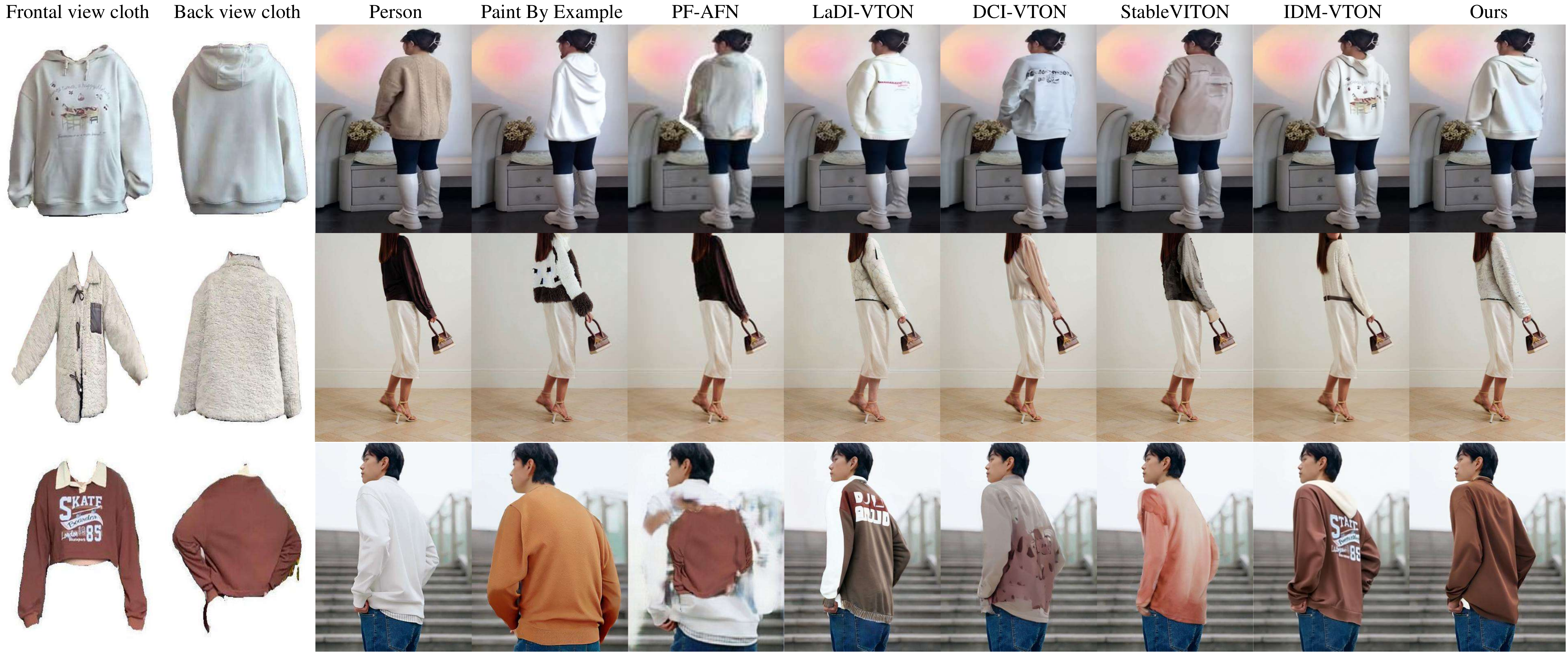}
    \vspace{-6mm}
    \caption{Comparison results on multi-view virtual try-on task.}
    \label{fig:sup_cmp_mv}
\end{figure*}

\noindent\textbf{Comparison Results on Multi-View VTON.}
In this section, we present more comparison results on the MVG dataset. Specifically, in the first row of Figure~\ref{fig:sup_cmp_mv}, due to the lack of adaptive selection of clothes, previous methods have difficulty in generating hoods of the original cloth. Moreover, in the second and third rows, previous methods often struggle to maintain fidelity to the original garments. In contrast, our method effectively addresses the aforementioned problems and generates high-fidelity results.

\begin{figure*}[t]
\centering
    \includegraphics[width=\textwidth]{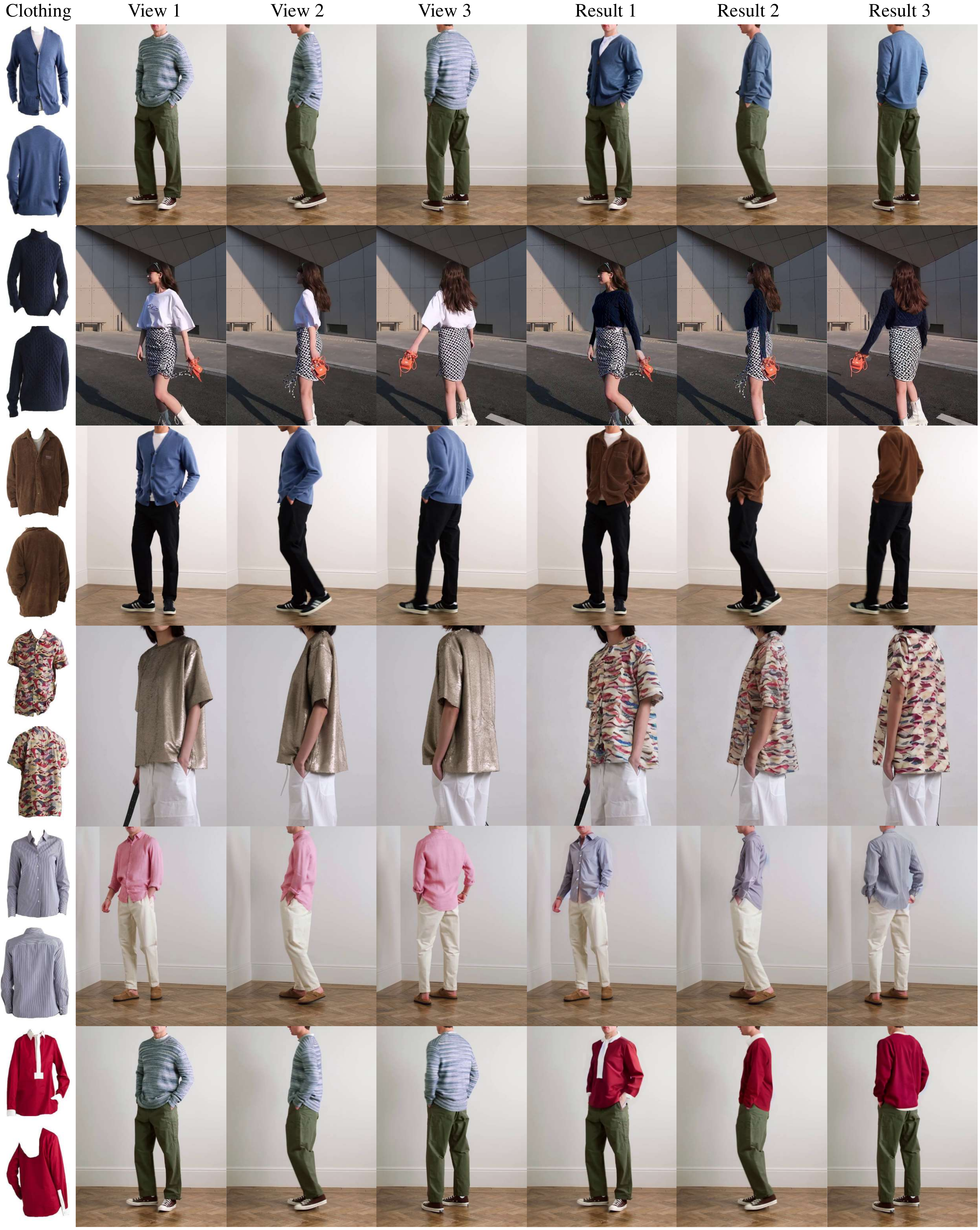}
    \vspace{-6mm}
    \caption{Multi-view results on multi-view virtual try-on task.}
    \label{fig:cmp_m1}
\end{figure*}

\noindent\textbf{Multi-View Results on Multi-View VTON.}
In this section, we present more multi-view results on the MVG dataset. Specifically, as shown in Figure \ref{fig:cmp_m1}, we show multiple groups of try-on results for the same person under different views, using the proposed method. In Figure \ref{fig:cmp_m1}, the first column displays frontal-view and back-view garments, the second to fourth columns depict persons from different views, while the fifth to seventh columns showcase the corresponding try-on results. As can be seen, our method can generate realistic dressing-up results of the multi-view person from the given two views of clothing. Furthermore, our method can retain the details well on the original clothing (\eg, the buttons in the fifth row) and generate high-fidelity try-on images even under occlusion (\eg, hair occlusion in the second row). In conclusion, the proposed method exhibits outstanding performance on multi-view virtual try-on task.

\begin{figure*}[t]
\centering
    \includegraphics[width=\textwidth]{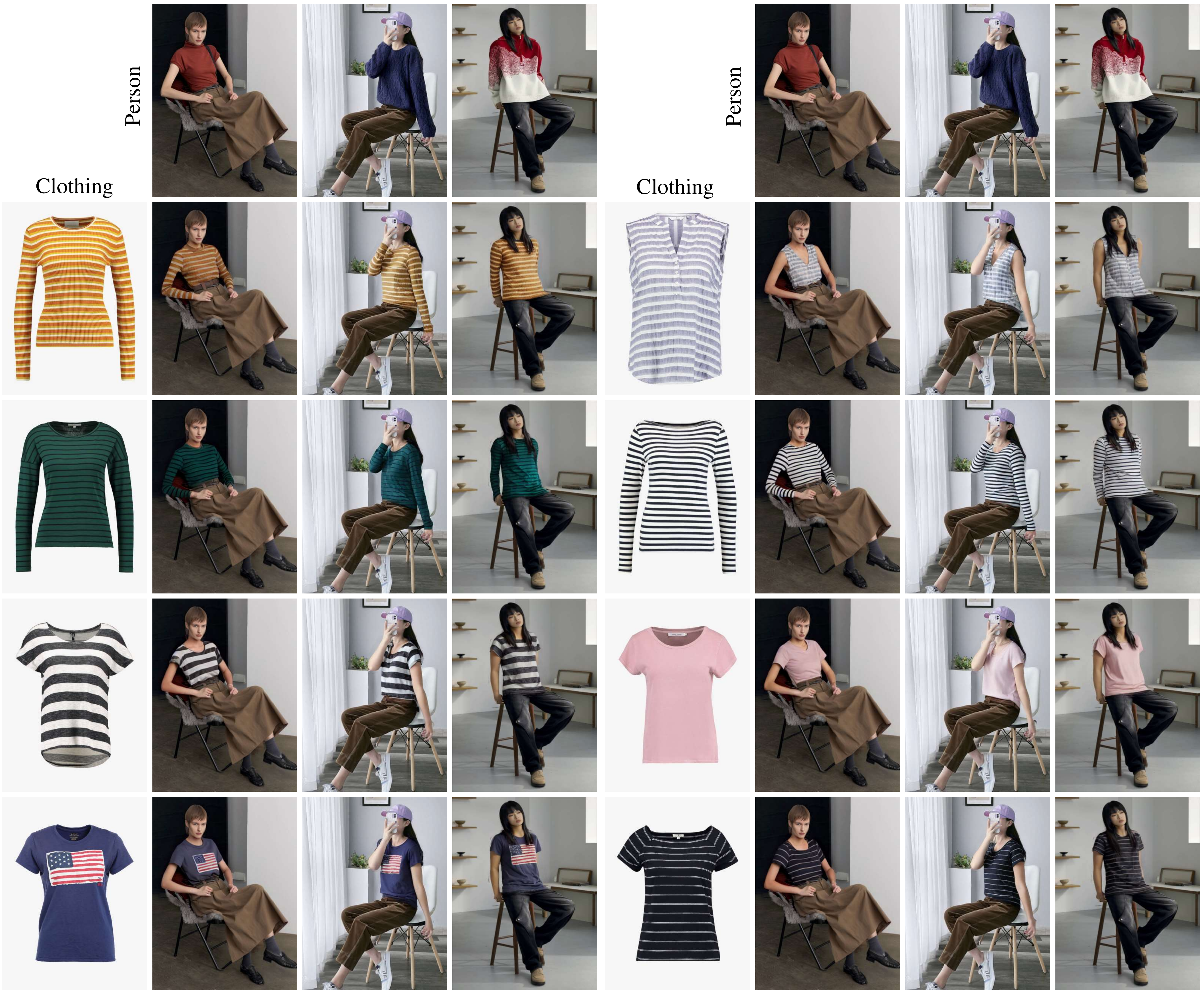}
    \vspace{-6mm}
    \caption{Complex human pose results on frontal-view virtual try-on task.}
    \label{fig:sup_complex_pose}
\end{figure*}

\noindent\textbf{Complex Human Pose Results on Frontal-View VTON.}
In this section, we provide more VTON results under complex human pose conditions in Figure~\ref{fig:sup_complex_pose}. It can be seen that our method can also generate high-fidelity try-on results even when the target person has a more complex pose.

\begin{figure*}[t]
\centering
    \includegraphics[width=\textwidth]{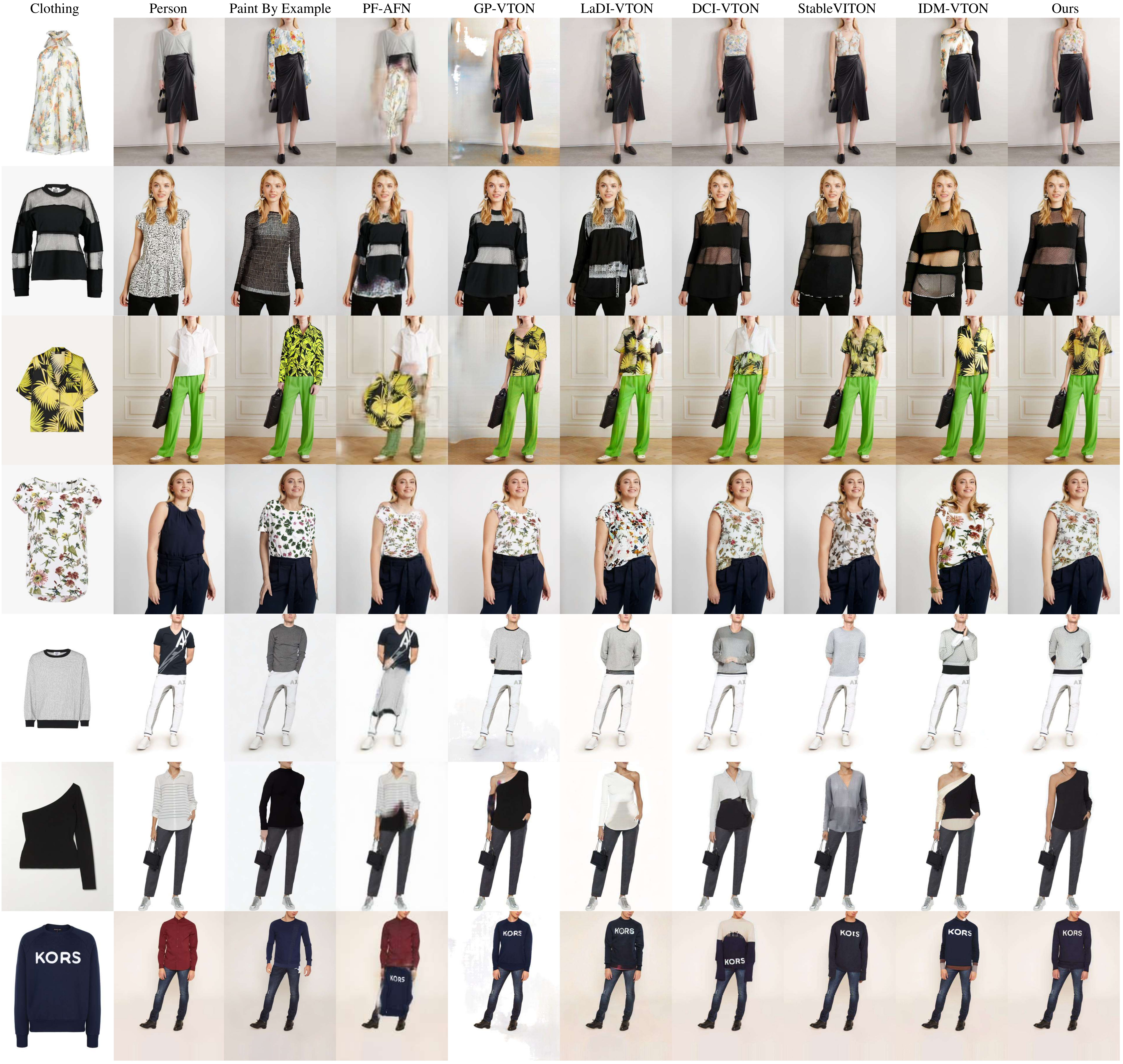}
    \vspace{-6mm}
    \caption{Qualitative comparisons on frontal-view virtual try-on task.}
    \label{fig:cmp_f2}
\end{figure*}

\noindent\textbf{Comparison Results on Frontal-View VTON.}
In this section, we show more visual comparison results on VITON-HD~\cite{choi2021viton} dataset and DressCode~\cite{morelli2022dress} dataset. The previous works include Paint By Example~\cite{yang2023paint}, PF-AFN~\cite{ge2021parser}, GP-VTON~\cite{xie2023gp}, LaDI-VTON~\cite{morelli2023ladi}, DCI-VTON~\cite{gou2023taming} and StableVITON~\cite{kim2023stableviton}. The results are shown in Figure \ref{fig:cmp_f2}. In the first and second row of Figure \ref{fig:cmp_f2}, it can be seen that our method better preserves the shape of the original clothing (\eg, the cuff in the second row), compared to the previous methods. In addition, our method outperforms previous methods in preserving high-frequency details, such as patterns on clothing in the fourth and sixth rows. Moreover, in contrast to previous methods, MV-VTON is not constrained by specific types of clothing and can achieve highly realistic effects across a wide range of garment styles (\eg, the garment in the third row and the collar in the eighth row). In summary, our method also has superiority on frontal-view virtual try-on task.

\begin{figure*}[t]
\centering
    \includegraphics[width=0.93\textwidth]{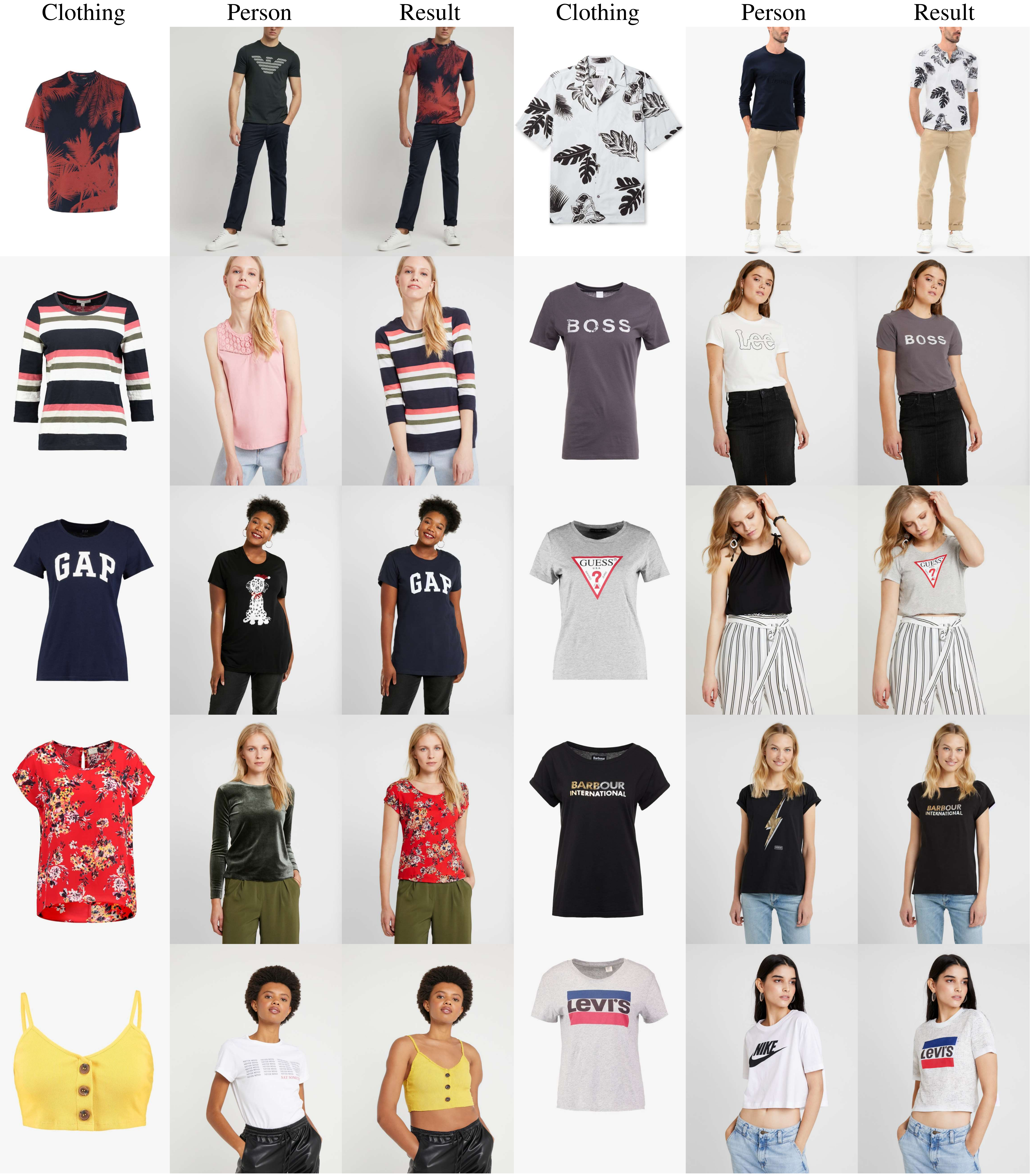}
    % \vspace{-7mm}
    \caption{Qualitative results of 1024$\times$768 resolution on frontal-view virtual try-on task.}
    \label{fig:cmp_f3}
\end{figure*}

\noindent\textbf{High Resolution Results on Frontal-View VTON.}
 In this section, we present more results at 1024$\times$768 resolution on VITON-HD~\cite{choi2021viton} and DressCode~\cite{morelli2022dress} datasets, as shown in the Figure~\ref{fig:cmp_f3}. Specifically, we utilize the model trained at 512$\times$384 resolution to directly test at 1024$\times$768 resolution. Despite the difference in resolutions between training and testing, our method can also produce high-fidelity try-on results. For instance, the generated images can preserve both the intricate patterns and text adorning the clothing (in the first row) while also effectively maintaining their original shapes (in the last row).

\begin{figure}[t!]
    \centering
    \includegraphics[width=0.475\textwidth]{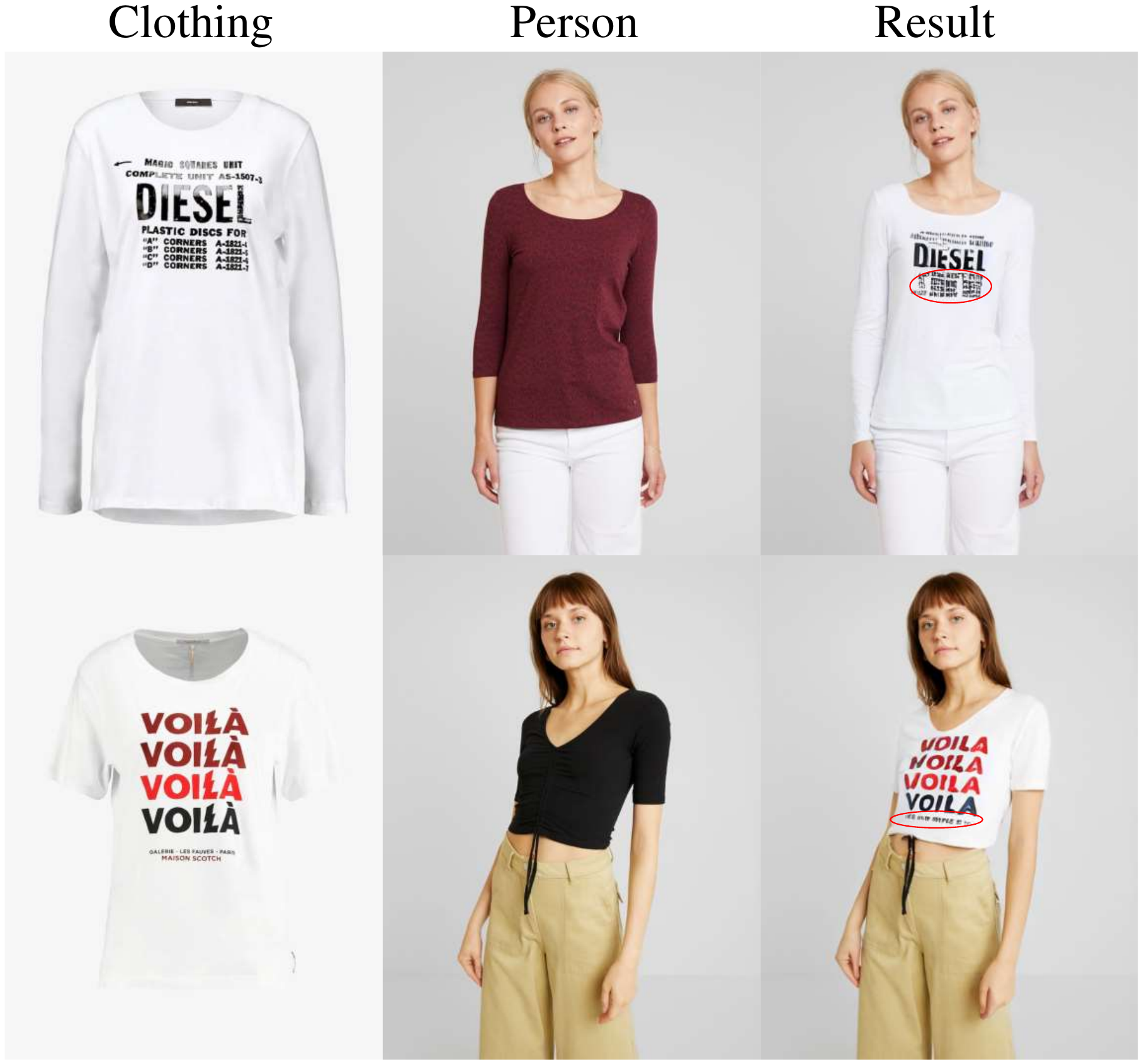}
    \vspace{-4mm}
    \caption{Visualization of bad cases on VITON-HD dataset.}
    \label{fig:limit}
\end{figure}

\section{LIMITATIONS}
Despite outperforming previous methods on both frontal-view and multi-view virtual try-on tasks, our method does not perform well in all cases. Figure~\ref{fig:limit} displays some unsatisfactory try-on results. As can be seen, although our method can preserve the shape and texture of original clothing (\eg, the 'DIESEL' text in the first row), it is difficult for it to fully preserve some smaller or more complex details (\eg, the parts circled in red). The reason for this phenomenon may be that these details are easily lost when inpainting in latent space. We will try to solve this issue in future work.

\clearpage

\bibliography{aaai25}

\end{document}